\documentclass[runningheads]{llncs}

\usepackage{eccv}

\usepackage{eccvabbrv}

\usepackage{graphicx}
\usepackage{booktabs}

\usepackage[accsupp]{axessibility}  

\usepackage{pifont}

\usepackage{amssymb}
\usepackage{mathtools}

\usepackage{float}
\usepackage{caption}

\usepackage{multirow}

\usepackage{microtype}
\usepackage{graphicx}
\usepackage{booktabs} 

\usepackage{hyperref}

\usepackage{orcidlink}

\usepackage{algorithmic}
\usepackage{algorithm}

\usepackage{bm}
\usepackage{mathtools}
\global\long\def\bx{\mathbf{\mathbf{x}}}%

\global\long\def\bx{\mathbf{x}}%
\global\long\def\bz{\mathbf{z}}%
\global\long\def\y{\mathrm{y}}%
\global\long\def\and{\cap}%
\usepackage{url}
\usepackage{amsmath}
\usepackage{gensymb}
\usepackage{booktabs}
\usepackage{multirow}
\usepackage{tabularray}
\usepackage{adjustbox}
\usepackage{subcaption}
\usepackage{pifont}
\usepackage{wrapfig}
\begin{document}

\title{Class-Prototype Conditional Diffusion Model with Gradient Projection for Continual Learning} 

\titlerunning{GPPDM}

\author{Khanh Doan\inst{1} \and
Quyen Tran\inst{1} \and
Tung Lam Tran\inst{1} \and
Tuan Nguyen\inst{2} \and
Dinh Phung\inst{2} \and
Trung Le\inst{2}}

\authorrunning{Doan et al.}

\institute{VinAI Research\\
\email{\{v.khanhdn10, v.quyentt15, v.lamtt12\}@vinai.io}\\
\and
Monash University\\
\email{\{tuan.ng, dinh.phung, trunglm\}@monash.edu}}

\maketitle

\begin{abstract}
Mitigating catastrophic forgetting is a key hurdle in continual learning. Deep Generative Replay (GR) provides techniques focused on generating samples from prior tasks to enhance the model's memory capabilities using generative AI models ranging from Generative Adversarial Networks (GANs) to the more recent Diffusion Models (DMs). A major issue is the deterioration in the quality of generated data compared to the original, as the generator continuously self-learns from its outputs. This degradation can lead to the potential risk of catastrophic forgetting (CF) occurring in the classifier. To address this, we propose the \textbf{G}radient \textbf{P}rojection Class-\textbf{P}rototype Conditional \textbf{D}iffusion \textbf{M}odel (GPPDM), a GR-based approach for continual learning that enhances image quality in generators and thus reduces the CF in classifiers. The cornerstone of GPPDM is a learnable class prototype that captures the core characteristics of images in a given class. This prototype, integrated into the diffusion model's denoising process, ensures the generation of high-quality images of the old tasks, hence reducing the risk of CF in classifiers. Moreover, to further mitigate the CF of diffusion models, we propose a gradient projection technique tailored for the cross-attention layer of diffusion models to maximally maintain and preserve the representations of old task data in the current task as close as possible to their representations when they first arrived. Our empirical studies on diverse datasets demonstrate that our proposed method significantly outperforms existing state-of-the-art models, highlighting its satisfactory ability to preserve image quality and enhance the model's memory retention.


\end{abstract}
\section{Introduction} \label{sec:introduction}
Continual Learning (CL) is a process where neural networks progressively acquire and accumulate knowledge from sequentially arriving tasks over time. Mimicking human learning, CL aims to develop models capable of swiftly adapting to evolving real-world environments. The primary challenge in CL is addressing \textit{Catastrophic Forgetting} (CF) \cite{french1999catastrophic}, which involves the model's need to learn new tasks without losing previously acquired knowledge, particularly when access to old data is not possible.

Continual learning strategies primarily focus on overcoming CF by preserving knowledge from previous tasks. This is particularly crucial in situations where privacy concerns prevent the retention of old raw data and the task ID remains unidentified during testing. In such cases, methods that employ generative models, specifically Generative Replay (GR), are notably effective for learning the distribution of past data. GR approaches \cite{Hanul2017CLDGR,kemker2018fearnet,gao23e_dmcl} usually adopt a two-part system comprising a generator and a classifier. The generator's role is to generate samples from prior tasks, which are then used to train the classifier. The classifier is designed to not only classify samples from the current task but also effectively categorize the generated samples, thereby helping to mitigate catastrophic forgetting. 

In the context of GR approaches, initial efforts utilized GAN and Variational Auto Encoder (VAE) architectures, which operate in a single-directional manner (from generator to classifier) and recycle generated samples, leading to a gradual decline in image quality. The recent DDGR method \cite{gao23e_dmcl} leverages the benefits of Diffusion Models (DMs) to more effectively prevent forgetting in generative models. DDGR employs a bi-directional interaction between the generator and the classifier. In the generator-to-classifier direction, the generator creates high-quality samples by denoising the noisy images with class information from previous tasks, which are then utilized to train the classifier. Conversely, in the classifier-to-generator direction, the classifier which is pretrained on previous tasks guides the generator in synthesizing high-quality samples from these tasks. However, our experimental studies revealed a noticeable decrease in the quality of generated images when the model was trained on a lengthy sequence of tasks, adversely affecting its capacity to retain acquired knowledge from earlier tasks.

To address the challenge of catastrophic forgetting in Continual Learning more efficiently, this paper introduces the \textbf{G}radient \textbf{P}rojection Class-\textbf{P}rototype conditional \textbf{D}iffusion \textbf{M}odel (GPPDM). Specifically, considering the bi-directional relationship operated under GR strategies as described above, in our model, the classifier aids the diffusion model in learning additional suitable prototypes. These prototypes will help preserve abstract information of old data, thus reducing the disadvantage of DDGR in gradually forgetting old knowledge. Consequently, when generating replay data in new tasks for training the classifier, the diffusion model then makes use of the information from both the labels and the corresponding learned prototypes to get a better quality of generated images. Moreover, after training a class $y$, this label and its prototype, which are inputs of the DM, will become invariant components. Therefore, to exploit learned prototypes more efficiently and further limit CF, we desire to preserve the corresponding knowledge on DM more effectively to maintain the representations of old task data in the current task as close as possible to their representations when they first arrived. Fortunately, this desideratum aligns with the strategy of GPM \cite{saha2021gradient}, which inspires us to design a novel approach to inject GPM into DM, specifically Cross-Attention layers. This approach is not only streamlined but also effective in avoiding forgetting effectively. In summary, our primary contributions can be outlined as follows:\\
    \begin{itemize}
    \item We propose an efficient training technique for the diffusion model by learning class prototypes, which capture the most representative examples of classes from previous tasks. This approach facilitates the generation of high-quality images through class-prototype conditional denoising. \\
    \item To further reduce forgetting, we suggest a novel design of GPM for DM. This approach creates an overall harmony when combined with learned class prototypes in preserving old knowledge effectively, under a simple implementation.\\
    \item The comprehensive results from benchmark datasets show that our proposed method significantly outperforms the current state-of-the-art model, achieving higher average accuracy and lower average forgetting rate.
    \end{itemize}

\section{Related Work}
\subsection{Continual Learning}

In general, CL techniques combating catastrophic forgetting are divided into three main groups: regularization-based, parameter isolation, and memory-based approaches.

\textbf{Regularization-based approaches} incorporate constraints to parameter updates \cite{zenke2017continual, aljundi2018memory} or enforce consistency with previous model outputs \cite{li2017learning, zhu2021prototype}.

\textbf{Parameter isolation approaches} aim to tackle the problem by assigning each task to a dedicated portion of the network, preventing the overwriting of weights and the loss of information. When no architectural size constraints apply, it is possible to grow new branches for new tasks while freezing previous task parameters \cite{10.5555/3326943.3327027, 8578908, pmlr-v80-serra18a}.  

\textbf{Memory-based approaches} exhibit superior performance across diverse settings, particularly excelling in scenarios where task boundaries during training and task IDs during testing are unspecified, providing increased flexibility. Particularly, these approaches utilize episodic memory to store past data \cite{robins1995catastrophic, chaudhry2018efficient, chaudhry2019continual, saha2021gradient}. 

\subsection{Generative Replay Continual Learning}
Generative replay (GR) leverages a generative model \cite{gan, vae, Jonathan2020DDPM} to create a replay memory for old tasks, mitigating catastrophic forgetting. Classical GR methods typically employ VAE \cite{vae} or GAN \cite{gan} as the generator for Continual Learning (CL) \cite{wu2018, achille2018, zhai2019lifelong, ramapuram2020lifelong, mundt2022unified}. For instance, DGR \cite{Hanul2017CLDGR} uses GAN to generate previous samples for data replaying. MeRGANs \cite{chenshen2018memory} addresses forgetting in GANs and excels in CL. Additionally, the mnemonics training framework proposed by \cite{liu2020mnemonics} generates optimizable exemplars. Building on the success of diffusion models \cite{Jonathan2020DDPM, ho2021classifierfree}, \cite{gao23e_dmcl} suggests using diffusion models as a generative replay for Class Incremental Learning (CIL). Despite achieving success in mitigating the catastrophic forgetting of classifiers in CL, existing generative replay approaches using VAE, GAN, and diffusion models still face catastrophic forgetting themselves, resulting in a reduction in the quality of generated images across tasks. This work proposes an efficient workaround to alleviate the catastrophic forgetting of a diffusion model-based generative replay. Key components of our approach include class-conditioned prototypes that summarize class information to guide the diffusion model and diversity exploration with neighbor mix style to enhance the diversity of generated images.

\section{Generative Replay CL with Diffusion Models}
\label{sec:background:continual_learning}
In what follows, we present the problem setting of task incremental CL and the technicality of generative replay CL with diffusion models.

\noindent\textbf{Problem Setting of Class Incremental CL:} In Class Incremental Learning (CIL), a seamless and continuous sequence of tasks, denoted as $\mathcal{D}^t = {(\bx^t_i, y^t_i)}_{i=1}^{N^t}$ for $t=1,\dots,T$, arrives in the system. Here, the image $\bx^t_i \in \mathbb{R}^{H \times W \times C}$ and the label $y^t_i \in \mathcal{Y}^t$ (the set of labels for task $t$). The goal is to train a deep neural network $f_\phi$ initialized as $f^0_\phi$. Upon arrival of each task $\mathcal{D}^t$, the model needs to adapt from $f^{t-1}_\phi$ to $f^t_\phi$, mainly relying on the current data in $\mathcal{D}^t$ without revisiting the data from previous tasks $\mathcal{D}^{1:t-1}$.
One of the most significant challenges in TIL is catastrophic forgetting, where the current model $f^t_\phi$ tends to easily forget information from the oldest tasks, hence leading to poor predictive performances on these tasks. 

\noindent\textbf{Generative Replay with Diffusion Models:} To address catastrophic forgetting, generative models like GAN \cite{gan} and VAE \cite{vae} are commonly employed to generate replay memory for old tasks \cite{Hanul2017CLDGR,kemker2018fearnet}. However, GANs often encounter the mode collapsing problem, and VAEs may generate blurry images \cite{gan_tutorial}. Recently, diffusion models \cite{sohl2015deep, ho2020denoising} have emerged as state-of-the-art generative models capable of producing high-quality and diverse images. To overcome the inherent issues of GANs/VAEs and capitalize on the advantages of diffusion models, \cite{gao23e_dmcl} proposes using diffusion models as a generative replay for TIL. Specifically, at the end of task $t$, it trains a diffusion model $\bm\epsilon^t_\theta(\bx_k, k)$ on the current data $\mathcal{D}^t$ and the replay memory of the previous tasks $\mathcal{M}^{1:t-1}$. This trained diffusion model $\bm\epsilon^t_\theta(\bx_k, k)$ is then employed to generate the replay memory $\mathcal{M}^{1:t}$ for the next task. 

\section{Our Proposed Approach} \label{sec:proposed_approach} 
\begin{figure}[t]
  \centering
     
    \includegraphics[width=0.8\linewidth]{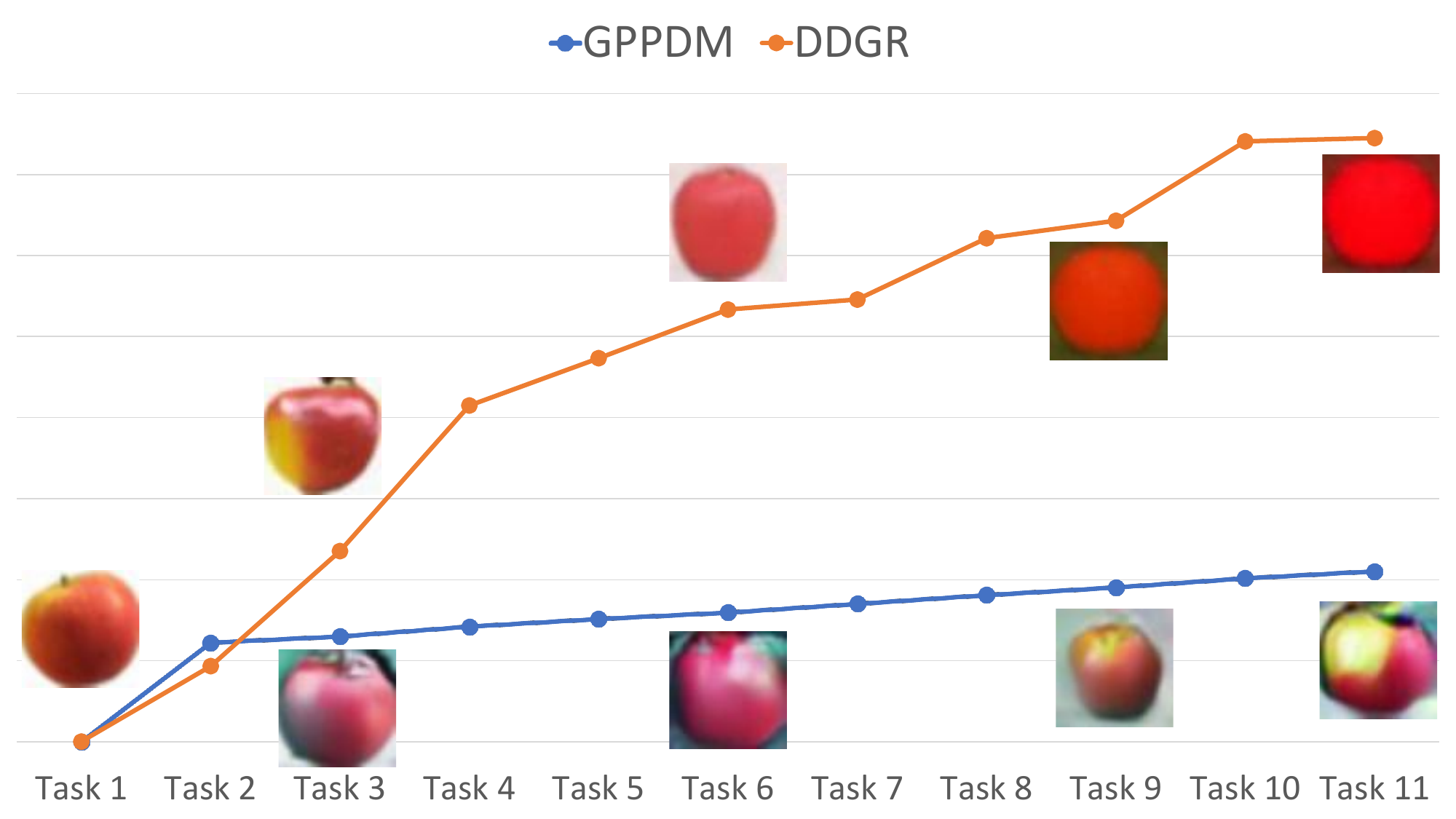}
   \vspace{-2mm}
   \caption{FID scores across tasks for generated images corresponding to the first task dataset of the baselines DDGR \cite{gao23e_dmcl} and our proposed GPPDM. The illustration belongs to the label class ``Apple'', whereas the leftmost one is the real photo. As can be observed, for DDGR, the quality of generated apples deteriorates significantly across tasks. Consequently, the generated apples of DDGR in some last tasks become hardly recognizable, while the generated apples of our proposed GPPDM maintain their quality more efficiently. Finally, FID scores across tasks for our GPPDM are significantly better than DDGR, especially for later tasks when the gap becomes more pronounced.}
   \label{fig:cl_fid}
   \vspace{-5mm}
\end{figure}
\subsection{Motivations}
Catastrophic Forgetting (CF) in classifiers poses a significant challenge in CIL. While applying generative models to replay old data can alleviate CF in classifiers, the generative models themselves may face CF within the context of CL. This occurs because, at the end of each task $t$, the generative model is trained based on the new task data $\mathcal{D}^t$ and the replay memory $\mathcal{M}^{1:t-1}$. The latter is generated from the generative model trained at the end of task $t-1$, using $\mathcal{D}^{t-1}$ and the previous replay memory $\mathcal{D}^{1:t-2}$. Notably, the quality of old data in the generative replay memory (e.g., $\mathcal{M}^1$) diminishes task by task, making it increasingly challenging for current and future diffusion models to maintain high-quality representations on this data.  

While it has been demonstrated in \cite{gao23e_dmcl} that Diffusion Models as generative replay can assist in mitigating the catastrophic forgetting of generative models in CL, our observations indicate that challenges still persist. As illustrated in Figure \ref{fig:cl_fid}, the quality of generated images for task 1's data significantly diminishes across tasks. Consequently, the generated images for some of the last tasks exhibit very low quality, making it difficult for the current model to recall old data and exacerbating the issue of catastrophic forgetting. 

Our primary approach to alleviate the CF of diffusion models involves learning class prototypes that succinctly summarize data within a given class. Subsequently, during both training and inference of new data, the diffusion models are conditioned on these class prototypes. It is worth noting that these class prototypes are learnable variables, which serve as a reserved information channel to effectively remind diffusion models of class concepts, contributing to reducing the CF across tasks. Moreover, to further mitigate the CF of diffusion models, we propose a gradient projection technique \cite{saha2021gradient} tailored for the cross-attention layer of diffusion models to maximally maintain and preserve the representations of old task data in the current task as close as possible to their representations when they first arrived.    



\subsection{Class-Prototype Conditional Diffusion Probabilistic Model}

When data  $(\bx, y) \in \mathcal{D}^t$ of task $t$ is arriving, 
thanks to the high-quality images in $\mathcal{D}^t$, we can train a diffusion model (DM) parameterized by $\theta$ to generate good images for task $t$. However, while learning a later task $t'>t$, to avoid forgetting old knowledge, we need to recall a small replay memory $\mathcal{M}^t$ of each task $t<t'$, which is generated from the DM being trained. It is worth noting that the quality of $\mathcal{M}^t$ gradually degrades, leading to the cumulative forgetting of data of class $y$ of task $t$ when trained and recalled in a future task $t'$.
To alleviate this CF and improve the quality of generated samples, we propose using additional learnable prototypes as an essential condition to preserve the most important and general information of old data. In particular, our diffusion model $p_{\theta}$ produces an image $\bx$ based on not only label $y$, but also the corresponding prototype $\mathbf{c}(y)$. For each label $y$, $\mathbf{c}(y)$ is an important factor that can capture and maintain abstract properties corresponding to data of this class. In our implementation, we treat it as a learnable factor and optimize it concurrently with $\theta$ during training. In what follows, we provide details of our class-prototype diffusion model.

\textbf{Forward process.} Given a predefined diffusion process of $L$ discrete timesteps, let $\bx_{0} \sim q(\bx_{0}|y)$ be clean data samples, where $q(\bx_0|y)$ represents the class-conditional data distribution over $\mathbb{R}^{d}$. Conditioned on $\bx_0$ and $y$, the joint distribution of the sequence $\bx_1, \bx_2, ..., \bx_L$ can be factorized using the chain rule of probability and the Markov property as follows:

\begin{equation}
    q(\bx_{1:L}|\bx_0,y) = \Pi_{l=1}^L q(\bx_l|\bx_{l-1},y).
\end{equation}

\textbf{Reverse process.} We begin the reverse process with a prior distribution $p(\bx_L|y, \mathbf{c}(y)) = \mathcal{N}(\bx_L; \mathbf{0},\mathbf{I})$, and then progressively denoise the noisy image $\bx_L$ to generate new data that belongs to class $y$. The process is encapsulated in the following equation:
\begin{equation}
    p_\theta(\bx_{0:L}|y, \mathbf{c}(y)) = p(\bx_L|y, \mathbf{c}(y))\Pi_{l=1}^L p_\theta(\bx_{l-1}|\bx_{l}, y, \mathbf{c}(y)).
\end{equation}

\textbf{Training objective.} Minimize the cross-entropy ($CE$) between $q\left(\bx_{0}|y\right)$ and $p_{\theta}\left(\bx_{0}|y, c(y)\right)$ to learn $p_\theta(\bx_{l-1}|\bx_{l}, y, \mathbf{c}(y))$:
\begin{align} & CE\left(q\left(\bx_{0}|y\right),p_{\theta}\left(\bx_{0}|y, \mathbf{c}(y)\right)\right)=-\mathbb{E}_{q\left(\bx_{0}|y\right)}\left[\log p_{\theta}\left(\bx_{0}|y, \mathbf{c}(y)\right)\right]\nonumber\\
 & \leq\mathbb{E}_{q\left(\bx_{0:L}|y\right)}\left[\log\frac{q\left(\bx_{1:L}\mid\bx_{0},y\right)}{p_{\theta}\left(\bx_{0:L}|y, \mathbf{c}(y)\right)}\right].
\end{align}
This leads to the optimization problem:
\begin{equation}
    \min_{\theta, \mathbf{c}(y)} \mathbb{E}_{q\left(\bx_{0:L}|y\right)}\left[\log\frac{q\left(\bx_{0:L}\mid \bx_{0},y\right)}{p_{\theta}\left(\bx_{0:L}|y, \mathbf{c}(y)\right)}\right].
\end{equation}

Using the $\epsilon$-network $\bm\epsilon_{\theta}\left(\bx_{l}, l, y, \mathbf{c}(y)\right)$ to predict the noise $\bm\epsilon$, we reach the following optimization problem for diffusion training: 

\begin{equation}\label{op:simplied_obj}
    \min_{\theta, \mathbf{c}(y)} \mathcal{L}_d \coloneqq \mathbb{E}_{\bm\epsilon,\bx_0,l,y} \left[\lVert \bm\epsilon - \bm\epsilon_{\theta}\left(\bx_{l},l, y,\mathbf{c}(y)\right)\rVert^2_2\right].
\end{equation}

Here we note that the class-prototype $\mathbf{c}(y)$ with $y \in \mathcal{Y}^t$ is trained during the task $t$ and kept fixed in the future tasks. 

\begin{figure*}[t]
    \begin{centering}
    \includegraphics[width=1.0\textwidth]{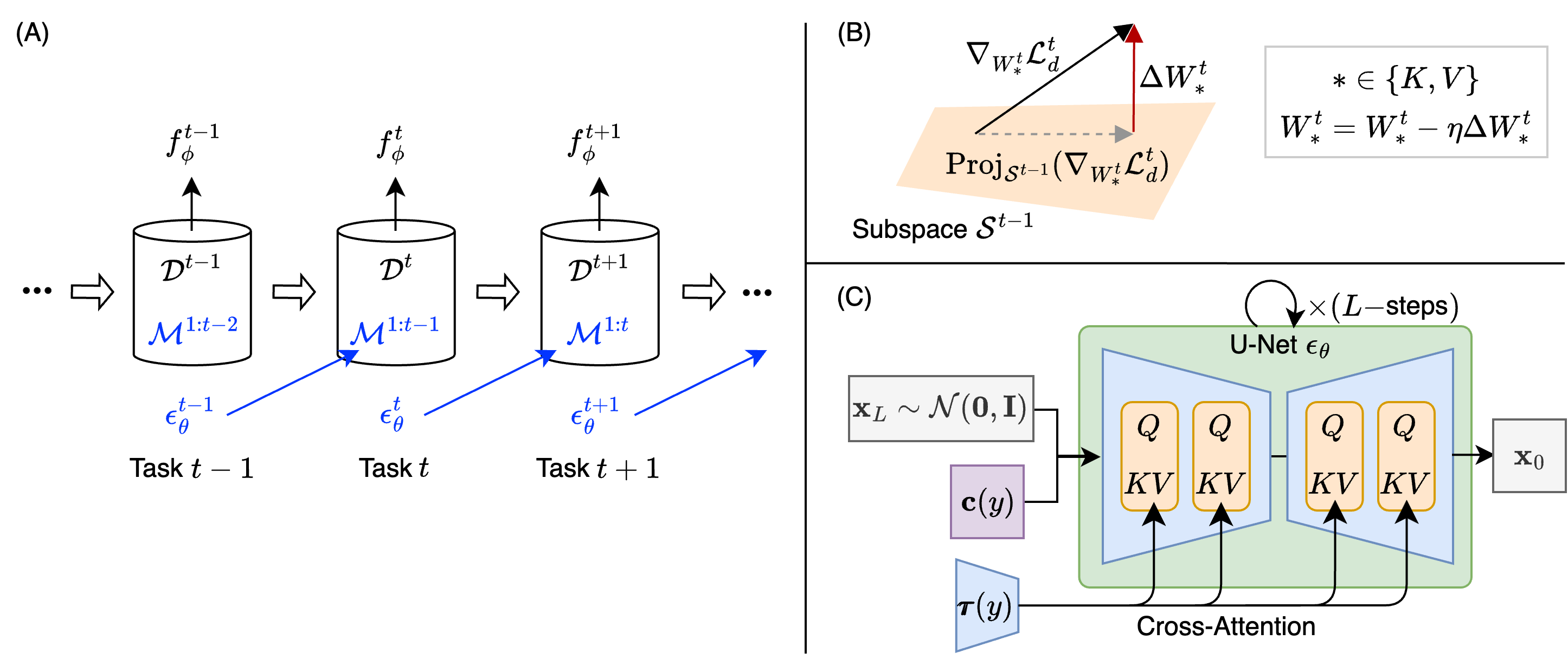}
    \par\end{centering}
    \caption{\textbf{(A) GPPDM framework for CL with generative replay}: During task $t$, the classifier $f^t_\phi$ trains on data from the current task $\mathcal{D}^t$ and on data from the previous task $\mathcal{M}^{1:t-1}$ generated by the diffusion model to reduce catastrophic forgetting. \textbf{(B) Diffusion training with Gradient Projection}: To maintain performance on images generated from previous tasks, the gradients of the diffusion loss $\mathcal{L}_d^t$ w.r.t. $W_K^t$ and $W_V^t$ are projected onto the subspace $\mathcal{S}^{t-1}$ and the orthogonal components, $\Delta W_K^t$ and $\Delta W_V^t$, are used to update $W_K^t$ and $W_V^t$ respectively. \textbf{(C) Diffusion sampling process}: The diffusion model begins by denoising the initial noise $\mathcal{N}(\mathbf{0}, \mathbf{I})$. Each denoising step incorporates class information from CLIP embeddings $\bm\tau(y)$ and the learned class prototype $\mathbf{c}(y)$. The class prototype is designed to capture the most distinctive features of a given class and assists the generator in synthesizing high-quality images.     \label{fig:Overall_framework-1}}
\end{figure*}

\begin{algorithm}[h]
\begin{algorithmic}[1]
    \REQUIRE Task $t$, dataset $\mathcal{D}^t$ with the label set $\mathcal{Y}^{t}$, classifier $f_\phi$, diffusion model $\bm\epsilon_{\theta}$, set of class-prototypes $\mathcal{C}$, pretrained CLIP model.
    \ENSURE Optimal $\phi^*$
    \FOR {$t=1,...,T$}
        \IF {$t\geq 2$}
            \STATE $\mathcal{M}^{1:t-1}=\textit{DiffusionSampler}(\mathcal{Y}^{1:t-1},\mathcal{C}^{1:t-1},\tilde{\bm\epsilon}_{\theta})$
            \STATE $(\bx,y) \sim \mathcal{D}^t \cup \mathcal{M}^{1:t-1}$
        \ELSE
            \STATE $(\bx,y) \sim \mathcal{D}^t$
        \ENDIF
        \STATE Update $\phi$ according to (\ref{eq:train_classifier}).
        \STATE Initialize $\mathbf{c}(y)$ according to (\ref{eq:init_c}).
        \STATE Update $\theta,\mathbf{c} \in \mathcal{C}^t$ according to (\ref{op:simplied_obj}) using the gradient projection technique.
    \ENDFOR
    \end{algorithmic}
    \caption{Pseudocode for training our GPPDM.\label{alg:training_GPPDM}}
\end{algorithm}

\begin{algorithm}[h]
    \begin{algorithmic}[1]
    \REQUIRE  Label set $\mathcal{Y}^{1:t-1}$, class-prototypes from previous tasks $\mathcal{C}^{1:t-1}$, classifier-free diffusion model $\tilde{\bm\epsilon}_{\theta}$, diffusion steps $L$, number of generated samples $R$.
    \ENSURE $\mathcal{M}^{t}$
    \STATE $\mathcal{M}^t = \varnothing$
    \FOR {$y \in \mathcal{Y}^{1:t-1}$}
        \STATE $r = 0$
        \WHILE {$r \leq R$}
            \STATE Get $\mathbf{c}(y)$ from $\mathcal{C}^{1:t-1}$
            \STATE $\bx_{L} \sim \mathcal{N}(\mathbf{0},\mathbf{I})$
            \FOR {$l=L,...,1$}
                \STATE $\bz \sim \mathcal{N}(\mathbf{0},\mathbf{I})$
                \STATE $\tilde{\bm\mu}_{\theta}=\frac{1}{\sqrt{\alpha_{l}}}\left(\bx_{l}-\frac{1-\alpha_{l}}{\sqrt{1-\bar{\alpha}_{l}}}\tilde{\bm\epsilon}_{\theta}\left(\bx_{l},l, \mathbf{c}(y),\bm\tau(y)\right)\right)$
                \STATE $\bx_{l-1} = \tilde{\bm\mu}_{\theta} + \sigma_{l} \bz$
            \ENDFOR
            \STATE $\mathcal{M}^t = \mathcal{M}^t \cup \{(\bx_0, y)\}$
            \STATE $r = r + 1$ 
        \ENDWHILE
    \ENDFOR 
    \end{algorithmic}
\caption{\textit{DiffusionSampler}\label{alg:diffusionsampler}}
    \end{algorithm}

\subsection{Gradient Projection for CL Diffusion Model} \label{gradient_projection}
Considering the CF phenomenon discussed above, it is desirable to further preserve and maintain the intermediate representations of data from task $t$ with respect to the latent diffusion model $\bm\epsilon_{\theta^{t'}}\left(\bx_{l},l, y,\mathbf{c}(y)\right)$ in future task $t'$, closely resembling those with respect to the diffusion model $\bm\epsilon_{\theta^t}\left(\bx_{l},l, y,\mathbf{c}(y)\right)$ in task $t$.          

This desideratum aligns with the gradient projection memory \cite{saha2021gradient} in the setting of Continual Learning (CL), where at each immediate layer, we need to compute a subspace which is the linear span of representations of the data from tasks so far, and then the model will be updated in the direction orthogonal to these subspaces. However, this approach is impractical in diffusion models due to (i) the large number of layers in the $\epsilon$-network diffusion model and (ii) the requirement to compute subspaces with respect to all $\bx_l, l=1,\dots, L$, given $\bx_0 = \bx$. In the following, we present how to achieve the above desideratum through a novel gradient projection technique tailored for the diffusion model in the context of image generation tasks.

For diffusion models \cite{Jonathan2020DDPM, rombach2022high}, the cross-attention layers where the label embeddings $\bm\tau_y \in \mathbb{R}^{d_\tau \times N}$ are injected to the model are crucially important. In this work, we tailor the cross-attention layers for our gradient projection technique. Specifically, the computation of a cross-attention layer is as follows:
\begin{equation}
Q^{t}=W_{Q}^{t}\varphi\left(\bx\right),K^{t}=W_{K}^{t}\boldsymbol{\tau}_{y},V^{t}=W_{V}^{t}\boldsymbol{\tau}_{y},
\end{equation}

where $\bx$ is a data example of task $t$, $\varphi(\bx) \in \mathbb{R}^{d_z \times M}$ is the latent representation inputted to the cross-attention layer, $W_Q^t \in \mathbb{R}^{d \times d_z}$, $W_K^t \in \mathbb{R}^{d \times d_\tau}$, and $W_V^t \in \mathbb{R}^{d \times d_\tau}$. Here, we use the superscript (e.g., $t$) to represent the model parameters and representations at a task (e.g., $t$). Eventually, the cross-attention is computed as $Attention\left(Q,K,V\right)=softmax\left(\frac{Q^{T}K}{\sqrt{d}}\right)V^T$.

Since the diffusion model performs best with the data of task $t$, at a future task $t'>t$, we aim to ensure that $V^{t'} \approx V^t$, $K^{t'} \approx K^t$, and $Q^{t'} \approx Q^t$ to maintain performance on the data of task $t$ when task $t'$ arrives. Enforcing the constraint $Q^{t'} \approx Q^t$ is particularly challenging due to the large cardinality of $\varphi(\bx)$, where $\bx$ can be any element in the sequence $\bx_0, \bx_1, \dots, \bx_L$ from the forward process of the diffusion model with $(\bx_0, y) \in \mathcal{D}^t$. This makes it impossible to construct a subspace for $\varphi(\bx)$. Therefore, to enforce this constraint, we simply update the parameter $W_Q^{t'}$ and other parameters to minimize the objective function in (\ref{op:simplied_obj}) with respect to the replay memory $\mathcal{M}^t$. 

To ensure $K^{t'} \approx K^t$, and $Q^{t'} \approx Q^t$, we can develop a gradient projection technique for updating $W_K^{t}$ and $W_V^{t}$. First, when task $t$ arrives, we compute the subspace $\mathcal{S}^t$ as the linear span of the column vectors of $\bm\tau(y)$, where $y \in \mathcal{Y}^t$. We then combine this subspace $\mathcal{S}^t$ with the current set of subspaces $\mathbf{S}^{t-1}$ to yield the set of subspaces up to task $t$: $\mathbf{S}^{t}$ (i.e., $\mathbf{S}^{t} = \mathbf{S}^{t-1} \cup \mathcal{S}^t$). To gain $W_K^{t}$ and $W_V^{t}$, we start from $W_K^{t} = W_K^{t-1}$ and $W_V^{t} = W_V^{t-1}$ and perform many update steps. Inspired by \cite{saha2021gradient}, at an update step, we denote $W_{K}^{t}=W_{K}^{t} - \eta\Delta W_{K}^t$ and $W_{V}^{t}=W_{V}^{t} - \eta\Delta W_{V}^t$ with the learning rate $\eta$, and require the row vectors in $\Delta W_{K}^t$ and $\Delta W_{V}^t$ to be orthogonal to $\mathbf{S}^{t-1}$ to preserve the performance of old classes before task $t$. Additionally, these orthogonal constraints can be conveniently implemented using the gradient projection technique. Specifically, let $\mathcal{L}^t_d$ be the diffusion loss in (\ref{op:simplied_obj}) at task $t$. We compute $\Delta W_{*}^{t}=\nabla_{W_{*}^{t}}\mathcal{L}_{d}^{t}-\text{Proj}_{\mathcal{S}^{t-1}}\left(\nabla_{W_{*}^{t}}\mathcal{L}_{d}^{t}\right)$ with $* \in \{K,V\}$ and use them to update $W_K^t$ and $W_V^t$ (see Figure \ref{fig:Overall_framework-1}).             

\subsection{Class-Prototype Conditional DM with Gradient Projection}
\label{sec:our_proposed_approach:class-prototype-conditional-dm-with-gr}
\phantom{abc}
\textbf{Training Methodology.} For each task $t$, we train the classifier $f^t_\phi$ by minimizing $CE$ loss w.r.t. $\phi$:
\begin{equation} \label{eq:train_classifier}
    \min_{\phi} \mathcal{L_C} \coloneqq \mathbb{E}_{(\bx,y)}\left[CE(f^t_\phi(\bx),y)\right],
\end{equation}
where $(\bx,y) \sim \mathcal{D}^t \cup \mathcal{M}^{1:t-1}$. Subsequently, the diffusion model $\bm\epsilon^t_{\theta}\left(\bx_{k},k, \mathbf{c}(y),y\right)$ learns data from $\mathcal{D}^t$ and replay memory $\mathcal{M}^{1:t-1}$ to generate samples for $\mathcal{M}^{1:t}$ that serves for the next task (as illustrated in Figure \ref{fig:Overall_framework-1}).

\textbf{Class-prototype Initialization and Update.} At task $t$, for each label $y^t_{i} \in \mathcal{Y}^t$, we initialize $\mathbf{c}(y^t_{i})$ with sample belonging to this label that our model $f^t_\phi$ has the highest confidence on prediction:
\begin{equation}\label{eq:init_c}
    \mathbf{c}(y^t_{i}) \coloneqq \text{argmax}_{\bx^t \in \mathcal{X}^{y^t_{i}}} p(y^t_{i}|\bx^t) = \text{argmin}_{\bx^t \in \mathcal{X}^{y^t_{i}}} \left[CE(f^t_\phi(\bx^t),y^t_{i})\right],
\end{equation}
where $\mathcal{X}^{y^t_{i}} \coloneqq \{\bx^t | (\bx^t, y^t) \in \mathcal{D}^t \text{ and } y^t = y^t_{i}\}$.

These class prototypes are then updated along with the diffusion model by minimizing the objective function in (\ref{op:simplied_obj}). Note that we apply the gradient projection technique as discussed in Section \ref{gradient_projection} to update $W_K^t$ and $W_V^t$ of the cross-attention layers. The detailed algorithms are outlined in Algorithms \ref{alg:training_GPPDM} and \ref{alg:diffusionsampler}.

\section{Experiments}
\label{sec:experiments}
We assess the effectiveness of our proposed GPPDM in the commonly encountered Class Incremental (CI) scenario \cite{van2018generative, van2019three} and Class Incremental with Repeatition (CIR) scenario \cite{cossu2021classincremental} in CL.

\subsection{Setup}
\label{sec:experiments:experimental_setup}
To ensure a balanced comparison, we adhere to the four experimental settings outlined in \cite{gao23e_dmcl}, applied to the CIFAR-100 and ImageNet \cite{deng2009imagenet} datasets for CI, and CORe50 \cite{lomonaco2017core50} for CIR. Besides, one more dataset is used, CUB-200 \cite{wah2011caltech}, to have a comprehensive view of the performance of GPPDM in CI.

\paragraph{Baselines.}
\label{sec:experiments:experimental_setup:baseline}
We compare our approach with the baselines including \textbf{Finetuning}, \textbf{SI} \cite{zenke2017continual}, \textbf{MAS} \cite{aljundi2018memory}, \textbf{EWC} \cite{kirkpatrick2017overcoming}, \textbf{IMM} \cite{lee2017overcoming}, \textbf{DGR} \cite{Hanul2017CLDGR}, \textbf{MeRGAN} \cite{chenshen2018memory}, \textbf{PASS} \cite{zhu2021prototype}, and \textbf{DDGR} \cite{gao23e_dmcl}. SI (Synaptic Intelligence), MAS (Memory Aware Synapses), EWC (Elastic Weight Consolidation), and IMM (Incremental Moment Matching) are methods that primarily focus on estimating the prior distribution of model parameters when assimilating new data. Generally, these approaches evaluate the significance of each parameter within a neural network, operating under the assumption of parameter independence for practicality. On the other hand, DGR (Deep Generative Replay) and MeRGAN leverage Generative Adversarial Networks (GANs) to create prior samples, facilitating data replay. Meanwhile, PASS represents a straightforward approach that does not rely on exemplars.

\paragraph{Datasets.}
\label{sec:experiments:experimental_setup:datasets}
For CI, we use two scenarios for each dataset CIFAR-100, ImageNet, and CUB-200. Half of the classes belong to the first task, the rest is divided equally among the remaining tasks. Additionally, CORe50 already has specific settings adapted to CIR. Specific details are as follows:
\begin{itemize}
    \item \textbf{CIFAR-100}: initial task consists of 50 classes with two cases of incremental tasks: 5 incremental tasks (i.e., 10 classes per task) and 10 incremental tasks (i.e., 5 classes per task). 
    \item \textbf{ImageNet}: initial task consists of 500 classes with two cases of incremental tasks: 5 incremental tasks (i.e., 100 classes per task) and  10 incremental tasks (50 classes per task). 
    \item \textbf{CUB-200}: initial task consists of 100 classes with two cases of incremental tasks: 5 incremental tasks (i.e., 20 classes per task) and  10 incremental tasks (10 classes per task). 
    \item \textbf{CORe50}: 79 batches as tasks are designed following \cite{lomonaco2017core50} from 50 classes (10 coarse labels, 5 subclasses for each label) in which 10 classes for the initial task and 5 for each subsequent one.
\end{itemize}
\begin{table*}[ht]
\small
\renewcommand{\arraystretch}{1.3}
\caption{Average accuracy and forgetting (\%) of methods across different architectures and numbers of classes in incremental tasks on CIFAR-100 and ImageNet. Underlined scores represent the second-best results.}
\vspace*{1mm}
\label{tab:main_table}
\centering
\resizebox{1.01\textwidth}{!} {
\begin{tabular}{c cc cc cc cc cc cc cc cc}
\hline 
& \multicolumn{8}{c}{ Average Accuracy $A_{T} (\uparrow)$ } & \multicolumn{8}{c}{ Average Forgetting $F_{T} (\downarrow)$ } \\
\hline
  & \multicolumn{4}{c}{CIFAR-100} & \multicolumn{4}{c}{ImageNet} & \multicolumn{4}{c}{CIFAR-100}  & \multicolumn{4}{c}{ImageNet} \\ 
  
  \cmidrule(lr){2-17}
  
  Method & \multicolumn{2}{c}{AlexNet} & \multicolumn{2}{c}{ResNet} & \multicolumn{2}{c}{AlexNet} & \multicolumn{2}{c}{ResNet} & \multicolumn{2}{c}{AlexNet} & \multicolumn{2}{c}{ResNet} & \multicolumn{2}{c}{AlexNet} & \multicolumn{2}{c}{ResNet} \\ 
  
  \cmidrule(lr){2-17} 
  & \textit{NC}=5 & 10 & 5 & 10 & 50 & 100 & 50 & 100 & 5 & 10 & 5 & 10 & 50 & 100 & 50 & 100 \\
  \hline
  Finetuning & 6.11 & 5.12 & 18.08 & 17.50 & 5.33 & 3.24 & 12.95 & 10.28 & 60.45 & 59.87 & 61.65 & 62.79 & 56.55 & 57.83 & 58.58 & 59.71\\
  
  SI & 16.96 & 13.57 & 26.45 & 23.15 & 19.38 & 14.38 & 28.88 & 24.38 & 48.58 & 50.18 & 52.27 & 56.65 & 41.18 & 45.94 & 41.93 & 44.56 \\
  
  EWC & 15.29 & 9.71 & 25.49 & 18.82 & 15.22 & 13.03 & 23.51 & 22.03 & 50.38 & 54.27 & 52.83 & 60.47 & 45.65 & 47.01 & 46.98 & 46.96 \\
  
  MAS & 20.13 & 18.94 & 29.94 & 28.28 & 16.35 & 14.51 & 31.25 & 25.51 & 45.75 & 45.37 & 49.38 & 51.31 & 44.85 & 45.70 & 39.55 & 44.34 \\
  
  IMM & 11.26 & 9.87 & 21.02 & 19.79 & 13.68 & 11.13 & 23.19 & 19.73 & 54.60 & 54.57 & 58.12 & 59.89 & 46.65 & 49.31 & 47.70 & 49.62 \\
  
  DGR & 42.49 & 38.16 & 52.96 & 48.94 & 43.94 & 38.81 & 53.32 & 47.56 & 24.08 & 26.52 & 26.36 & 31.14 & 17.31 & 22.52 & 17.96 & 21.84 \\
  
  MeRGAN & 46.03 & 43.23 & 57.19 & 55.69 & — & — & — & — & 36.95 & 26.49 & 20.12 & 22.55 & — & — & — & —\\
  
  PASS & 53.21 & 48.65 & 62.30 & \underline{60.63} & — & — & — & — & 27.35 & 19.43 & 16.97 & 21.21 & — & — & — & —\\
  
  DDGR & \underline{59.20} & \underline{52.22} & \underline{63.40} & 60.04 & \underline{53.86} & \underline{52.21} & \underline{64.83} & \underline{61.26} & \underline{23.00} & \underline{16.86} & \underline{15.34} & \underline{19.25} & \underline{6.98} & \underline{7.82} & \underline{5.65} & \underline{7.73}\\

 \hline
 \textbf{GPPDM} & \textbf{76.35} & \textbf{62.76} & \textbf{77.79} & \textbf{68.07} & \textbf{58.49} & \textbf{52.71} & \textbf{72.88} & \textbf{65.98} & \textbf{8.46} & \textbf{9.63} & \textbf{6.47} & \textbf{6.21} & \textbf{5.94} & \textbf{4.40} & \textbf{4.30} & \textbf{4.75}\\

  \hline

\end{tabular}}
\vspace{-3mm}
\end{table*}

\paragraph{Models.}
\label{sec:experiments:experimental_setup:models}
Following previous works \cite{rebuffi2017icarl, liu2020mnemonics, de2021continual}, three model architectures are used for the classifier: AlexNet \cite{krizhevsky2012imagenet}, 32-layer ResNet \cite{he2016deep}, and Vision Transformer ViT-B/16 \cite{dosovitskiy2020image}. Similarly, following \cite{ho2020denoising}, popular UNet architecture is used for the diffusion model.

\paragraph{Evaluations.}
\label{sec:experiments:experimental_setup:evaluations}
For CI evaluations, two primary metrics are employed: Average Accuracy and Average Forgetting, with their specifics available in the supplementary materials. In the context of CIR, the test set consists of a comprehensive dataset encompassing all classes, evaluated through sequential tasks. Here, the performance of the current model on this dataset is measured in terms of Average Accuracy. 


\subsection{Experimental Results}
\label{sec:experiments:results}


\paragraph{Results in CI scenario.}
Table \ref{tab:main_table} displays the outperformance of our GPPDM compared to baseline models on the CIFAR-100 and ImageNet datasets, under different architectures and number of classes per subsequent task (NC). On CIFAR-100, GPPDM consistently beats the second best method by a large margin in terms of average accuracy and average forgetting, especially on a longer sequence of tasks. For example, with AlexNet, GPPDM improves the average accuracy of DDGR by around 17\% and 10\% when $NC=5$ and $10$, respectively.
On a more challenging ImageNet dataset, a similar pattern can be observed. In the case of AlexNet with $NC=100$, despite a slight improvement in regard to average accuracy, our method shows its ability to clearly reduce average forgetting from $7.82\%$ to $4.40\%$. Moreover, Table \ref{tab:fair_and_more} shows that our GPPDM also surpasses the main baseline DDGR on CUB-200.

\paragraph{Results in CIR scenario.} A test dataset is maintained throughout the training process, therefore it is reasonable to consider maximum average accuracy instead of the forgetting term. As presented in \cite{gao23e_dmcl}, DDGR performs well in this scenario, the most obvious example is sample quality visualized, which is as good as real data. In experiments with AlexNet architecture, our results are the same as DDGR. But with ResNet, we reach outperformance: $8\%$ for $A_{T}$ and $11\%$ for $\max_{1 \leq i \leq T}(A_{T})$, see Table \ref{tab:cir_results}.

\begin{table}
\centering
\label{tab:ablation}
\vspace*{-5mm}
\caption{Comparison in Average accuracy (\%) between our GPPDM and DDGR for CIR scenario with CORe50 dataset.\label{tab:cir_results}}
\begin{tabular}{c cc cc}
\hline
               & \multicolumn{2}{c}{$A_{T} (\uparrow)$} & \multicolumn{2}{c}{$\max_{1 \leq i \leq T}(A_{i}) (\uparrow)$} \\\hline
Method         & AlexNet            & ResNet            & AlexNet                        & ResNet                        \\\hline
DDGR           & $31.00$    & $40.66$           & $31.00$                      & $42.00$                     \\
\textbf{GPPDM} & \textbf{31.05}            & \textbf{48.69}    & \textbf{34.30}                 & \textbf{53.31}               \\\hline
\end{tabular}
\vspace*{-10mm}
\end{table}

\subsection{Ablation Study}
\label{sec:experiments:ablation_study}

\begin{figure}[t]
  \centering
  
    \includegraphics[width=0.8\linewidth]{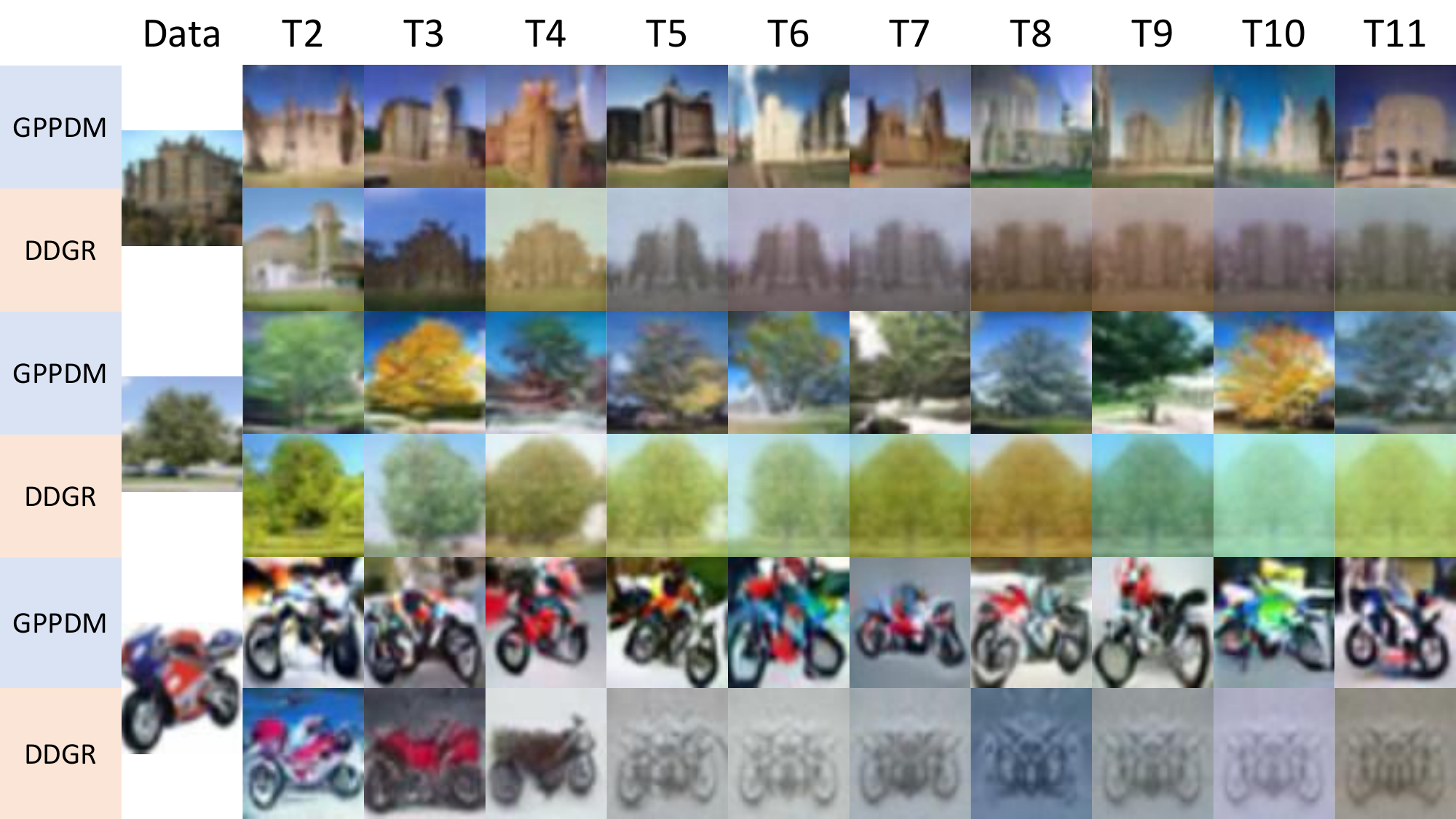}
   \caption{Comparison of generated images from GPPDM and DDGR on CIFAR-100}\label{fig:cf_of_GPPDM}
\end{figure}

\begin{figure}[t]
  \centering
   \includegraphics[width=1.0\linewidth]{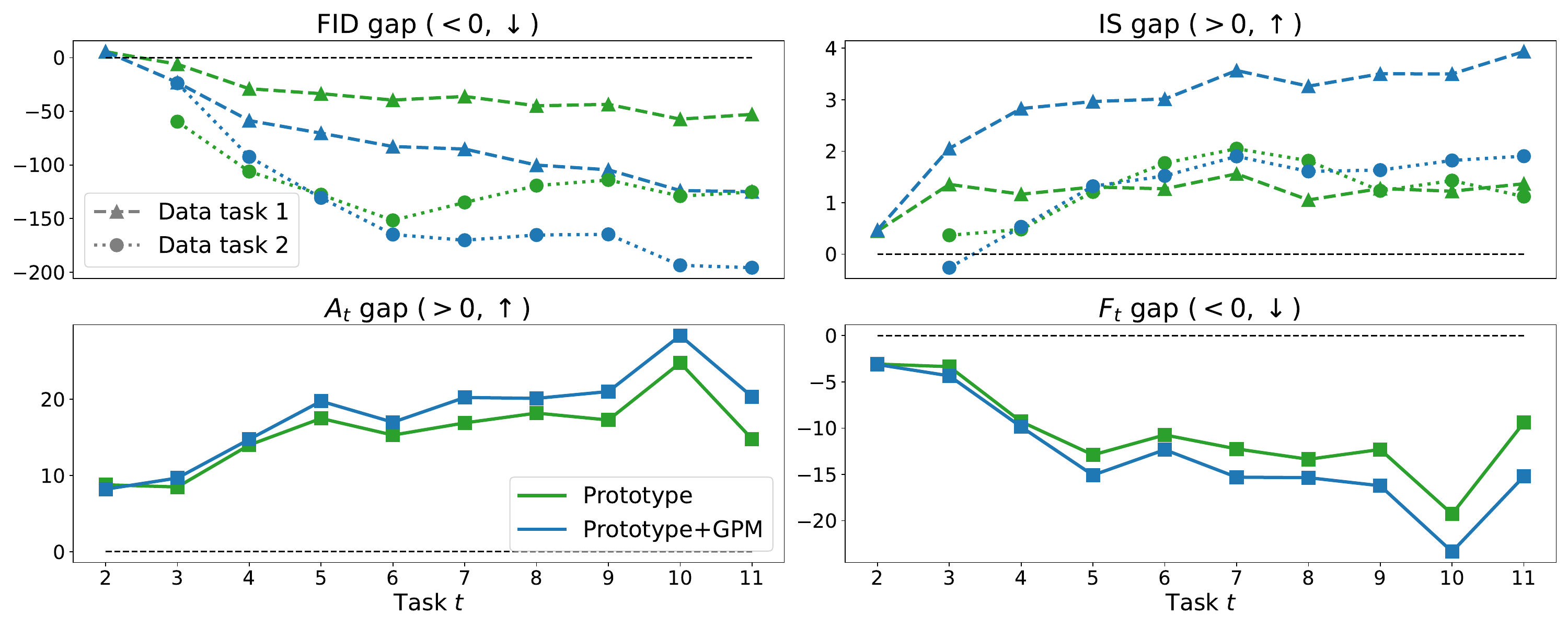}
   \caption{Outperformance of our methods compare to DDGR during sequential tasks in settings of CIFAR-100 ($NC=5$) with AlexNet is presented by gaps in Fréchet inception distance (FID), Inception Score (IS), Average Accuracy ($A_{t}$), and Average Forgetting ($F_{t}$). }\label{fig:compare}
   \vspace*{-5mm}
\end{figure}

\paragraph{Forgetting comparison in DMs.}
Although DDGR demonstrates an effective ability to preserve sample quality across numerous tasks in CIR settings, it cannot maintain this performance in CI settings, where each class's real data appears only once across tasks. Figure \ref{fig:cf_of_GPPDM} clearly shows that DDGR faces severe generation catastrophic forgetting, as seen in later tasks where its generated images become increasingly blurry and difficult to recognize. In contrast, our GPPDM effectively maintains key object features, demonstrating a more efficient approach to mitigating generational catastrophic forgetting. This is evident in the clarity and recognizability of the images generated by GPPDM, even in successive tasks. Figure \ref{fig:compare} provides a more detailed look at the gaps across tasks in some metrics regarding image quality as well as the performance differences they bring. The gaps are computed as the differences in the results of DDGR and two variants of our GPPDM. Data belonging to task 1 and task 2 are selected to evaluate sample quality (FID between synthetic and real data, and IS of synthetic data). Dashed lines and triangles are used to describe for data task 1, dotted lines and circles are used for data task 2.


\begin{wrapfigure}{r}{0.48\columnwidth}
  \centering  
   \includegraphics[width=0.39\textwidth]{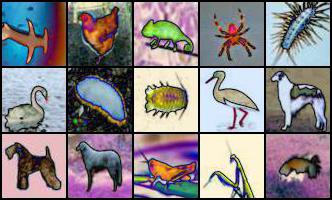}
   \caption{Visualization of prototypes.}\label{fig:learn_class_prototype}
\end{wrapfigure}

\paragraph{Learning Class-Prototypes.}Figure \ref{fig:learn_class_prototype} presents the class-prototypes achieved after training. Notably, these prototypes tend to emphasize and focus more intensely on the objects present in the initial images. Such a concentration on object-specific information is clearly advantageous, guiding our GPPDM to preserve class-concept information effectively in subsequent tasks.


\paragraph{Contribution of each proposed component.} As shown in Table \ref{tab:proposal_term}, when solely using either class-prototype guidance or tailored gradient projection technique, the forgetting of DDGR is significantly mitigated, resulting in higher average accuracy and lower average forgetting. Gradient projection seems to be better at maintaining stability of the model than prototype, but when the two are combined, the average forgetting is further reduced by around 1\% across most settings. 

\begin{table}
\centering
\label{tab:ablation}
\vspace*{-5mm}
\caption{Contribution of each proposed component on CIFAR-100.\label{tab:proposal_term}}
\begin{tabular}{c cc cc cc cc}
\hline
                        & \multicolumn{4}{c}{$A_T (\uparrow)$}                     & \multicolumn{4}{c}{$F_T (\downarrow)$}                   \\\hline
\multirow{2}{*}{Method} & \multicolumn{2}{c}{AlexNet} & \multicolumn{2}{c}{ResNet} & \multicolumn{2}{c}{AlexNet} & \multicolumn{2}{c}{ResNet} \\\cmidrule(lr){2-9}
                        & $NC=5$        & 10          & 5            & 10          & 5            & 10           & 5            & 10          \\\hline
Default (DDGR)          & 59.20         & 52.22       & 63.40        & 60.04       & 23.00        & 16.86        & 15.34        & 19.25        \\
Prototype               & 70.89         & 61.01       & 76.56        & 66.52       & 13.97        & 11.71        & 10.79        & 8.73        \\
GPM                     & 75.81         & 61.98       & 77.64        & 67.68       & 9.04         & 10.78        & 7.53         & 7.41        \\
Prototype + GPM         & \textbf{76.35}         & \textbf{62.76}       & \textbf{77.79}        & \textbf{68.07}       & \textbf{8.46}         & \textbf{9.63}         & \textbf{6.47}         & \textbf{6.21}    \\\hline   
\end{tabular}
\vspace*{-5mm}
\end{table}

\paragraph{The memory cost of GPPDM.} 
In comparison with DDGR, GPPDM requires saving a prototype per class and, which is equivalent to saving only one image. Nonetheless, as previously shown, these prototypes can notably improve DDGR. Here, in Table \ref{tab:fair_and_more}, we present a fairer comparison by equipping DDGR a buffer with one real image per old class to train the diffusion model and the classifier. Generally, replaying one sample is helpful but can not enhance DDGR's performance as much as ours. This is because our learned prototypes can capture class concepts better than the saved images, and is also due to the effectiveness of our cross-attention gradient projection.
\begin{table}
\centering
\caption{A fair comparison to DDGR in which a real image for each class is chosen to put in the generative replay memory (i.e., DDGR + buffer). \label{tab:fair_and_more}}
\begin{tabular}{c c cc cc cc cc cc}
\hline
\multirow{3}{*}{Metric}             & \multirow{3}{*}{Method} & \multicolumn{6}{c}{CIFAR-100}                                                                       & \multicolumn{4}{c}{CUB-200}                           \\\cmidrule(lr){3-12}
                                    &                         & \multicolumn{2}{c}{AlexNet}     & \multicolumn{2}{c}{ResNet}      & \multicolumn{2}{c}{ViT}         & \multicolumn{2}{c}{AlexNet} & \multicolumn{2}{c}{ViT} \\\cmidrule(lr){3-12}
                                    &                         & 5              & 10             & 5              & 10             & 5              & 10             & 10       & 20               & 10     & 20             \\\hline
\multirow{3}{*}{$A_T (\uparrow)$}   & DDGR                    & 59.20          & 52.22          & 63.40          & 60.04          & 91.77          & 88.51          & 23.42    & 20.43            & 30.39  & 29.38          \\
                                    & DDGR+buffer             & 62.82          & 55.22          & 65.68          & 61.73          & 93.27          & 90.69          & 24.19         & 22.64            & 35.35       & 29.09          \\\cmidrule(lr){2-12}
                                    & \textbf{GPPDM}          & \textbf{76.35} & \textbf{62.76} & \textbf{77.79} & \textbf{68.07} & \textbf{93.66} & \textbf{92.29} & \textbf{27.52}    & \textbf{30.10}   & \textbf{39.92}  & \textbf{34.69} \\\hline
\multirow{3}{*}{$F_T (\downarrow)$} & DDGR                    & 23.00          & 16.86          & 15.34          & 19.25          & 6.17           & 7.36           & 17.20    & 9.19             & 17.46  & 10.00          \\
                                    & DDGR+buffer             & 21.29          & 16.85          & 19.03          & 9.94           & 4.80           & 5.49           & 8.11         & 6.00             & 10.91       & 9.73           \\\cmidrule(lr){2-12}
                                    & \textbf{GPPDM}          & \textbf{8.46}  & \textbf{9.63}  & \textbf{6.47}  & \textbf{6.21}  & \textbf{4.49}  & \textbf{4.29}  & \textbf{1.94}     & \textbf{4.52}    & \textbf{8.36}   & \textbf{7.33} \\\hline
\end{tabular}
\end{table}

\section{Conclusion}
Catastrophic forgetting poses a critical challenge in the realm of continual learning. Generative replay, a method employing a generative model to recreate a replay memory for old tasks, aims to reinforce the classifier's understanding of past concepts. However, generative models themselves may encounter catastrophic forgetting, impeding the production of high-quality old data across tasks. Recent solutions, such as DDGR, integrating diffusion models, aim to reduce this issue in generation, yet they still face significant challenges. To address this, in this paper, we propose the Class-Prototype conditional Diffusion Model with Gradient Projection (GPPDM) to notably enhance the mitigation of generation catastrophic forgetting. The core idea involves the acquisition of a class-prototype that succinctly encapsulates the characteristics of each class. These class-prototypes then serve as guiding cues for diffusion models, acting as a mechanism to prompt their memory when generating images for earlier tasks. Through empirical experiments on real-world datasets, our GPPDM demonstrates its superiority, significantly outperforming current leading methods.

\clearpage  

%
%
\bibliographystyle{splncs04}
\bibliography{ref}

\newpage

\title{\textbf{Supplementary Material for\\``Class-Prototype Conditional Diffusion Model\\ with Gradient Projection for Continual Learning''}}


\titlerunning{GPPDM}


\authorrunning{Doan et al.}




\maketitle

\section{Training objective of Conditional Diffusion Probabilistic Model} \label{sec:training_obj}
To learn $p_\theta(\bx_{l-1}|\bx_{l},y,\mathbf{c}(y))$, we apply the variational approach, which involves minimizing the cross-entropy divergence between $q(\bx_0|y)$ and $p_\theta(\bx_0|y,\mathbf{c}(y))$, represented by:

\begin{align}
    &CE\left(q\left(\bx_{0}|y\right),p_{\theta}\left(\bx_{0}|y,\mathbf{c}(y)\right)\right)\\
    &=-\mathbb{E}_{q\left(\bx_{0}|y\right)}\left[\log p_{\theta}\left(\bx_{0}|y,\mathbf{c}(y)\right)\right]\\
    &=-\mathbb{E}_{q\left(\bx_{0}|y\right)}\left[\int q\left(\bx_{1:L}\mid \bx_{0}, y\right)\log p_{\theta}\left(\bx_{0}|y,\mathbf{c}(y)\right)d\bx_{1:L}\right]\\
    &=-\mathbb{E}_{q\left(\bx_{0}|y\right)}\left[\int q\left(\bx_{1:L}\mid \bx_{0},y\right)\log\frac{p_{\theta}\left(\bx_{0:L}|y,\mathbf{c}(y)\right)}{p_{\theta}\left(\bx_{1:L}\mid \bx_{0},y,\mathbf{c}(y)\right)}d\bx_{1:L}\right]\\
    &=-\mathbb{E}_{q\left(\bx_{0}|y\right)}\left[\int q\left(\bx_{1:L}\mid \bx_{0},y\right)\log\frac{p_{\theta}\left(\bx_{0:L}|y,\mathbf{c}(y)\right)}{q\left(\bx_{1:L}\mid \bx_{0},y\right)}\frac{q\left(\bx_{1:L}\mid \bx_{0},y\right)}{p_{\theta}\left(\bx_{1:L}\mid \bx_{0},y,\mathbf{c}(y)\right)}d\bx_{1:L}\right]\\
    &=-\mathbb{E}_{q\left(\bx_{0}|y\right)}\Big[\int q\left(\bx_{1:L}\mid \bx_{0},y\right)\log\frac{p_{\theta}\left(\bx_{0:L}|y,\mathbf{c}(y)\right)}{q\left(\bx_{1:L}\mid \bx_{0},y\right)}d\bx_{1:L}\nonumber\\
    &\phantom{:=}+D_{KL}\left(q\left(\bx_{1:L}\mid \bx_{0},y\right)\Vert p_{\theta}\left(\bx_{1:L}\mid \bx_{0},y,\mathbf{c}(y)\right)\right)\Big]\\
    &\leq -\mathbb{E}_{q\left(\bx_{0}|y\right)}\left[\int q\left(\bx_{1:L}\mid \bx_{0},y\right)\log\frac{p_{\theta}\left(\bx_{0:L}|y,\mathbf{c}(y)\right)}{q\left(\bx_{1:L}\mid \bx_{0},y\right)}d\bx_{1:L}\right]\\
    &=-\mathbb{E}_{q\left(\bx_{0:L}|y\right)}\left[\log\frac{p_{\theta}\left(\bx_{0:L}|y,\mathbf{c}(y)\right)}{q\left(\bx_{1:L}\mid \bx_{0},y\right)}\right]\\
    &=\mathbb{E}_{q\left(\bx_{0:L}|y\right)}\left[\log\frac{q\left(\bx_{1:L}\mid \bx_{0},y\right)}{p_{\theta}\left(\bx_{0:L}|y,\mathbf{c}(y)\right)}\right].
\end{align}

The optimization problem then becomes:
\begin{equation}
    \min_{\theta} \mathcal{L} \coloneqq \mathbb{E}_{q\left(\bx_{0:L}|y\right)}\left[\log\frac{q\left(\bx_{1:L}\mid \bx_{0},y\right)}{p_{\theta}\left(\bx_{0:L}|y,\mathbf{c}(y)\right)}\right].
\end{equation}

We further derive the objective function of interest as:
\begin{align}
    \log\frac{q\left(\bx_{1:L}\mid \bx_{0},y\right)}{p_{\theta}\left(\bx_{0:L}|y,\mathbf{c}(y)\right)}
    &=\log \frac{q\left(\bx_{L}\mid \bx_{0},y\right)}{p\left(\bx_{L},y,\mathbf{c}(y)\right)} \times \prod_{l=2}^{L}\frac{q\left(\bx_{l-1}\mid \bx_{l},\bx_{0}, y\right)}{p_{\theta}\left(\bx_{l-1}\mid \bx_{l},y,\mathbf{c}(y)\right)}\nonumber\\
    &\phantom{:=}\times\frac{1}{p_{\theta}\left(\bx_{0}\mid \bx_{1},y,\mathbf{c}(y)\right)}\\
    &=\log\frac{q\left(\bx_{L}\mid \bx_{0},y\right)}{p\left(\bx_{L},y,\mathbf{c}(y)\right)} + \sum_{l=2}^{L}\log\frac{q\left(\bx_{l-1}\mid \bx_{l},\bx_{0},y\right)}{p_{\theta}\left(\bx_{l-1}\mid \bx_{l},y,\mathbf{c}(y)\right)}\nonumber\\
    &\phantom{:=}-\log p_\theta(\bx_0|\bx_1,y,\mathbf{c}(y)),
\end{align}
leading us to the following conclusion:
\begin{align}
    \mathcal{L} &= \mathbb{E}_{q}\left[\log\frac{q\left(\bx_{L}\mid \bx_{0},y\right)}{p\left(\bx_{L}|y,\mathbf{c}(y)\right)} + \sum_{l=2}^{L}\log\frac{q\left(\bx_{l-1}\mid \bx_{l},\bx_{0},y\right)}{p_{\theta}\left(\bx_{l-1}\mid \bx_{l},y,\mathbf{c}(y)\right)}-\log p_\theta(\bx_0|\bx_1,y,\mathbf{c}(y))\right]\\
    &=\mathbb{E}_{q}\Big[\underbrace{D_{KL}\left(q\left(\bx_{L}\mid\bx_{0},y\right)\|p\left(\bx_{L}|y,\mathbf{c}(y)\right)\right)}_{\mathcal{L}_{L}}\nonumber\\
    &\phantom{:=}+ \sum_{l>1}\underbrace{D_{KL}\left(q\left(\bx_{l-1}\mid\bx_{l},\bx_{0},y\right)\|p_{\theta}\left(\bx_{l-1}\mid\bx_{l},y,\mathbf{c}(y)\right)\right)}_{\mathcal{L}_{l-1}}\nonumber\\
    &\phantom{:=}\underbrace{-\log p_{\theta}\left(\bx_{0}\mid\bx_{1},y,\mathbf{c}(y)\right)}_{\mathcal{L}_{0}}\Big]
\end{align}
\section{Implementation Specification and Additional Experimental Results}
\label{sec:additional_exp}

\subsection{Additional experiments}
\label{sec:additional_exp:additional_experiment}
\begin{figure}[t]
  \centering
   \includegraphics[width=0.7\textwidth]{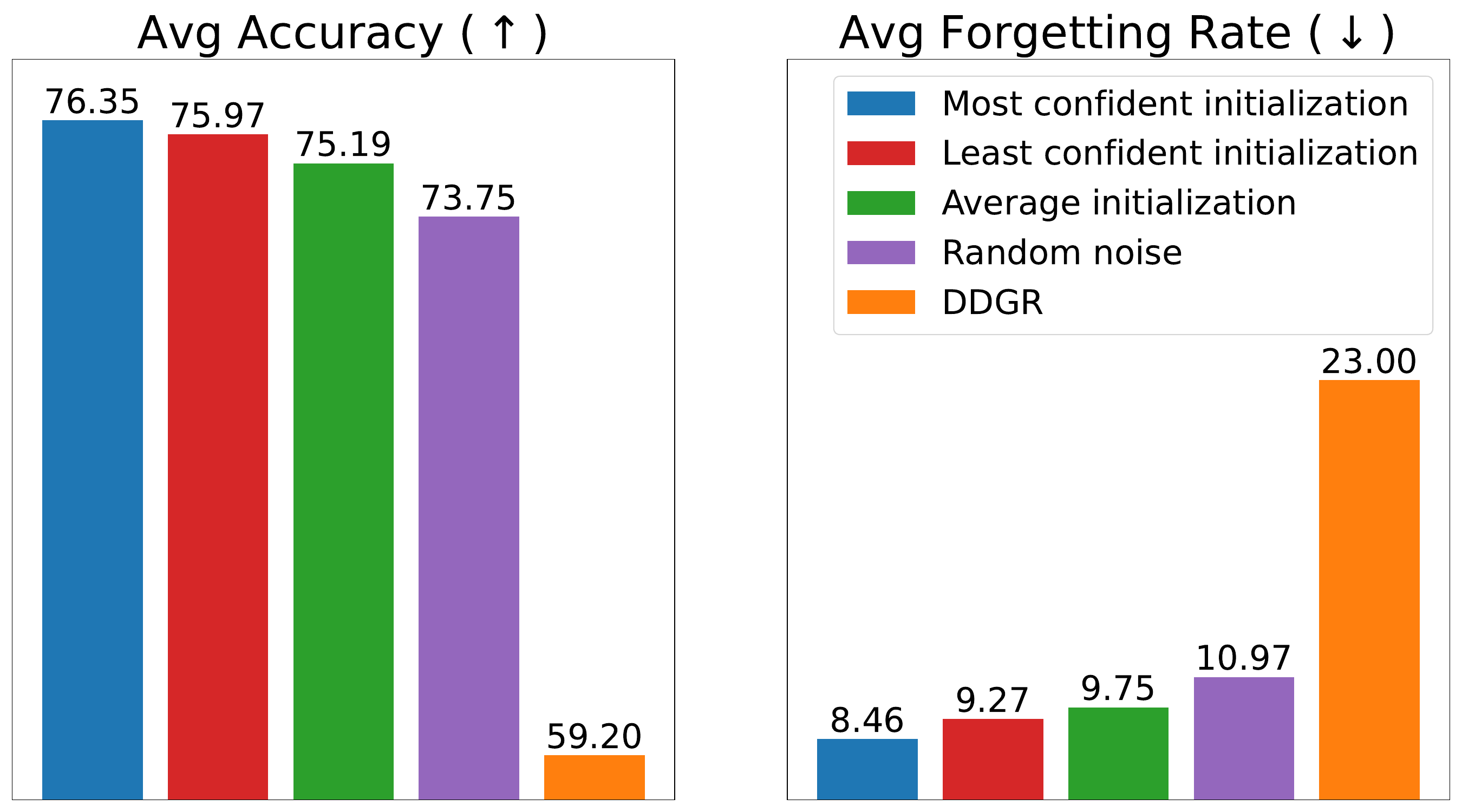}

   \caption{Comparison of final average accuracy $A_{T} (\uparrow)$ and final average forgetting $F_{T} (\downarrow)$ in the setting of CIFAR-100 $(NC=5)$, AlexNet with different class-prototype initialization strategies relative to DDGR.}
   \label{fig:initialization}
\end{figure}

\paragraph{Analysis of Class-Prototype Initialization.}
In the CIFAR-100 ($NC=5$) experiment using the AlexNet model, we examine various initializations of class-prototypes. In addition to the initialization used in the main paper, given $y^t_{i} \in \mathcal{Y}^t$, three alternatives are also investigated as follows:
\begin{itemize}
    \item Most confident initialization (as presented in Section \ref{sec:our_proposed_approach:class-prototype-conditional-dm-with-gr})
    \begin{equation}\label{eq:init_c}
    \mathbf{c}(y^t_{i}) \coloneqq \arg\max_{\bx^t \in \mathcal{X}^{y^t_{i}}} p(y^t_{i}|\bx^t) = \arg\min_{\bx^t \in \mathcal{X}^{y^t_{i}}} \left[CE(f^t_\phi(\bx^t),y^t_{i})\right].
    \end{equation}
    \item Least confident initialization\
    \begin{equation}\label{eq:init_c}
    \mathbf{c}(y^t_{i}) \coloneqq \arg\min_{\bx^t \in \mathcal{X}^{y^t_{i}}} p(y^t_{i}|\bx^t) = \arg\max_{\bx^t \in \mathcal{X}^{y^t_{i}}} \left[CE(f^t_\phi(\bx^t),y^t_{i})\right].
    \end{equation}
    \item Average initialization in the same class
    \begin{equation}
    \mathbf{c}(y^t_{i}) \coloneqq \frac{1}{|\mathcal{X}^{y^t_{i}}|}\sum_{\bx^t \in \mathcal{X}^{y^t_{i}}}{\bx^t}.
    \end{equation}
    \item Random initialization
    \begin{equation}
        \mathbf{c}(y) \sim \mathcal{N}(\mathbf{0}, \mathbf{I}).
    \end{equation}
\end{itemize}
Figure \ref{fig:initialization} illustrates the implicit relationship between the generator and the classifier in how to initialize the class prototypes, revealing that the most confident initialization method yields the best performance. Averaging samples in pixel space appears to be less effective, leading to worse outcomes compared to even the least confident initialization approach. Additionally, while the random initialization method results in slightly poorer performance, it significantly outperforms the baseline (DDGR).

\paragraph{Experiment without a large first task.}
Table \ref{tab:equality_num_cls_avg_acc} presents the Average Accuracy after each task on CIFAR-100 where all tasks have the same number of classes ($20$) with AlexNet as classifier architecture. Our method yields better results compared to DDGR and PASS across all tasks. However the gap of the final Average Accuracy $(6.31)$ is not as much as in other cases which have a larger first task $(NC=5:\text{ }17.15,\text{ } NC=10:\text{ }10.54)$. There are two reasons: fewer tasks are considered, and the capacity of the first task is smaller. Hence, it becomes simpler for the classifier to remember old tasks.

\begin{table}[]
\centering
\caption{Average Accuracy $A_{i} (\uparrow)$ across tasks on CIFAR-100 $5$ tasks, $20$ classes / task, AlexNet model.
\label{tab:equality_num_cls_avg_acc}}
\begin{tabular}{c ccccc}
\hline
              & \multicolumn{5}{c}{Average Accuracy}            \\
              \hline
Method        & Task 1 $(A_{1})$ & Task 2 $(A_{2})$ & Task 3 $(A_{3})$ & Task 4 $(A_{4})$ & Task 5 $(A_{5})$ \\
\hline
PASS          & 70.51   & 52.17   & 50.41   & 47.39   & 44.32   \\
DDGR          & \underline{70.65}   & \underline{54.33}   & \underline{53.35}   & \underline{51.16}   & \underline{49.03}   \\
\hline
GPPDM & \textbf{70.90}   & \textbf{66.30}   & \textbf{61.10}   & \textbf{56.79}   & \textbf{55.34}  \\
\hline
\end{tabular}
\end{table}

\begin{figure}[t]
  \centering
   \includegraphics[width=0.7\textwidth]{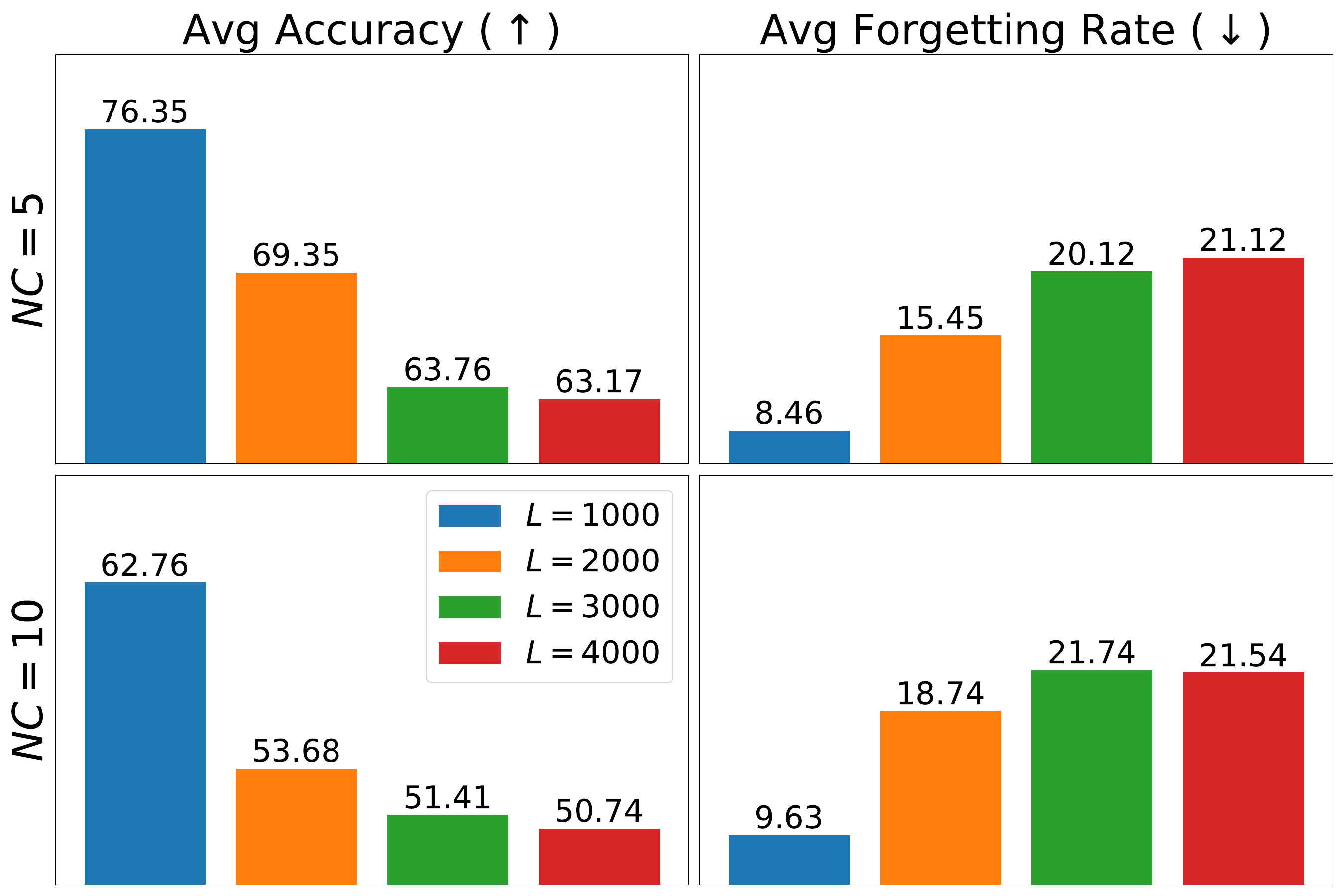}

   \caption{Results on CIFAR-100, AlexNet model, while changing the number of diffusion timesteps.}
   \label{fig:diffusion_timesteps}
\end{figure}

\paragraph{The effect of diffusion steps.}
Increasing the number of diffusion timesteps $(K)$ in training (from $1000$ to $2000$, $3000$, $4000$) while keeping the number of training steps and the number of inference timesteps the same $(100)$ makes the generator converge more slowly and makes it more difficult to form an image, thereby making the results worse. Figure \ref{fig:diffusion_timesteps} shows final Average Accuracy and final Average Forgetting of AlexNet model for CIFAR-100 dataset in two cases $NC=5$ and $NC=10$.

\subsection{Architecture / Hyperparameters}
\label{sec:additional_exp:hyperparam}

\paragraph{Data preparation and preprocessing.}
\label{sec:additional_exp:data_preparation}
In our implementation, which is grounded on the DDGR framework as delineated in \cite{gao23e_dmcl}, a standard preprocessing protocol is employed across three distinct datasets: CIFAR-100, ImageNet, CUB-200, and CORe50 \cite{lomonaco2017core50}. Central to this protocol is the resizing of image samples to a uniform dimension of $64 \times 64$. However, some exceptions are made: Vision Transformers (ViT-B/16) only get inputs with size of $224 \times 224$; the rest case is for ImageNet and CUB-200, wherein the $64 \times 64$ resized samples undergo an additional resizing step, being scaled up to $256 \times 256$ before being fed to the classifier (AlexNet and ResNet32). This resizing process, applied consistently across all datasets, employs a bilinear technique, ensuring uniformity in sample processing.

\begin{figure}[t]
  \centering
   \includegraphics[height=\textheight]{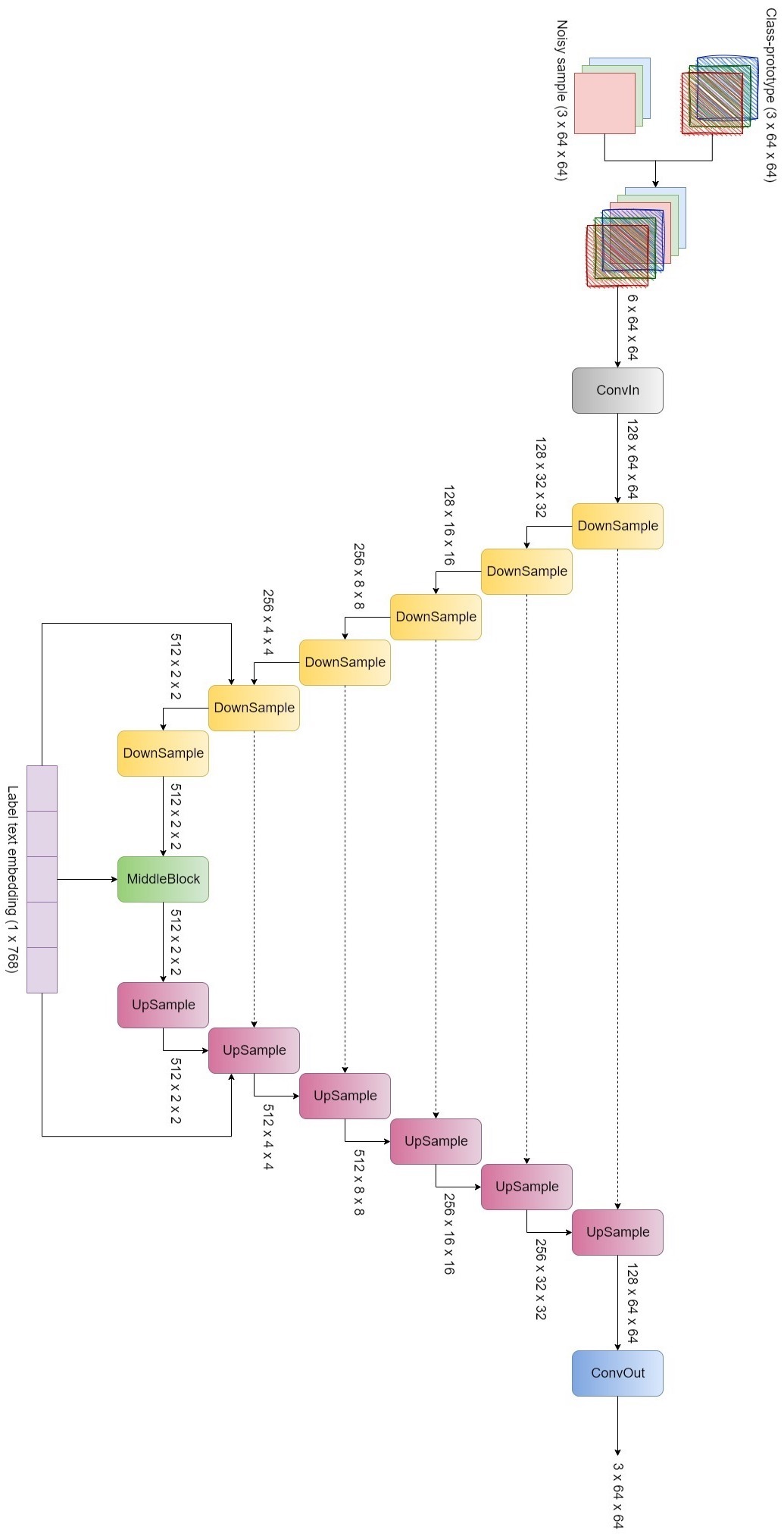}

   \caption{UNet diffusion model.}
   \label{fig:unet}
\end{figure}

\paragraph{Architecture.}
We utilize UNet architecture in our implementation. Specifically, we acquire a noisy sample denoted as $\bx_{l}$ at time step $l$, which mirrors the dimensions of real data, measuring $3 \times 64 \times 64.$ Simultaneously, our class-prototype, $\mathbf{c}(y),$ shares this spatial dimensionality. To facilitate seamless integration into our subsequent processes, we concatenate $\bx_{l}$ and $\mathbf{c}(y)$ in the channel dimension, resulting in a composite tensor measuring $6 \times 64 \times 64.$ This composite tensor is then seamlessly integrated into the UNet model for further processing and analysis. Label text embedding, output of CLIP model, having dimensions of $1 \times 768$, is used as key and value of Cross Attention operation in some specific blocks. Figure \ref{fig:unet} presents the pipeline of UNet model.

\paragraph{Label text embeddings.}
In this section, we detail the method of using text embedding for labels using CLIP.

\begin{itemize}
    \item CI scenario: Datasets used in this scenario are CIFAR-100, CUB-200, and ImageNet, which have a relatively clear distinction between the classes. Creating label text embedding is as simple as follows:

\begin{align}
    \y \mapsto \bm\tau(y) = \mathbf{CLIP}(``\text{a photo of a }[\mathbf{classname}(y)]")
\end{align}

    \item CIR scenario: The CORe50 dataset has a little difference in the meaning of the classes compared to the above two sets. This dataset has a hierarchical labels set, in which there are 10 coarse labels, each includes 5 fine labels. Therefore, label text embedding follows the following formula:

\begin{align}
    &\y \mapsto \bm\tau(y) = \mathbf{CLIP}(``\text{a photo of a } [\mathbf{classname}(y)] \nonumber\\
    &\phantom{===========}\text{ in category } [\mathbf{classfineindex}(y)]")
\end{align}

where $\mathbf{classname}(\cdot)$ and $\mathbf{classfineindex}(\cdot)$ are presented in the Table \ref{tab:core50_label}

\end{itemize}
\begin{table}[]
\centering
\caption{Details CORe50 dataset labels. \label{tab:core50_label}}
\vspace*{-3mm}
\resizebox{0.66\textwidth}{!} {
\begin{tabular}{c cccc}
\cmidrule(lr){1-5}
Coarse label                     & Label index $(y)$ & Fine label       & Class name     & Class fine index \\\cmidrule(lr){1-5}
\multirow{5}{*}{plug\_adapter}   & 0                 & plug\_adapter1   & plug adapter   & 1                \\\cmidrule(lr){2-5}
                                 & 1                 & plug\_adapter2   & plug adapter   & 2                \\\cmidrule(lr){2-5}
                                 & 2                 & plug\_adapter3   & plug adapter   & 3                \\\cmidrule(lr){2-5}
                                 & 3                 & plug\_adapter4   & plug adapter   & 4                \\\cmidrule(lr){2-5}
                                 & 4                 & plug\_adapter5   & plug adapter   & 5                \\\cmidrule(lr){1-5}
\multirow{5}{*}{mobile\_phone}   & 5                 & mobile\_phone1   & mobile phone   & 1                \\\cmidrule(lr){2-5}
                                 & 6                 & mobile\_phone2   & mobile phone   & 2                \\\cmidrule(lr){2-5}
                                 & 7                 & mobile\_phone3   & mobile phone   & 3                \\\cmidrule(lr){2-5}
                                 & 8                 & mobile\_phone4   & mobile phone   & 4                \\\cmidrule(lr){2-5}
                                 & 9                 & mobile\_phone5   & mobile phone   & 5                \\\cmidrule(lr){1-5}
\multirow{5}{*}{scissor}         & 10                & scissor1         & scissor        & 1                \\\cmidrule(lr){2-5}
                                 & 11                & scissor2         & scissor        & 2                \\\cmidrule(lr){2-5}
                                 & 12                & scissor3         & scissor        & 3                \\\cmidrule(lr){2-5}
                                 & 13                & scissor4         & scissor        & 4                \\\cmidrule(lr){2-5}
                                 & 14                & scissor5         & scissor        & 5                \\\cmidrule(lr){1-5}
\multirow{5}{*}{light\_bulb}     & 15                & light\_bulb1     & light bulb     & 1                \\\cmidrule(lr){2-5}
                                 & 16                & light\_bulb2     & light bulb     & 2                \\\cmidrule(lr){2-5}
                                 & 17                & light\_bulb3     & light bulb     & 3                \\\cmidrule(lr){2-5}
                                 & 18                & light\_bulb4     & light bulb     & 4                \\\cmidrule(lr){2-5}
                                 & 19                & light\_bulb5     & light bulb     & 5                \\\cmidrule(lr){1-5}
\multirow{5}{*}{can}             & 20                & can1             & can            & 1                \\\cmidrule(lr){2-5}
                                 & 21                & can2             & can            & 2                \\\cmidrule(lr){2-5}
                                 & 22                & can3             & can            & 3                \\\cmidrule(lr){2-5}
                                 & 23                & can4             & can            & 4                \\\cmidrule(lr){2-5}
                                 & 24                & can5             & can            & 5                \\\cmidrule(lr){1-5}
\multirow{5}{*}{glass}           & 25                & glass1           & glass          & 1                \\\cmidrule(lr){2-5}
                                 & 26                & glass2           & glass          & 2                \\\cmidrule(lr){2-5}
                                 & 27                & glass3           & glass          & 3                \\\cmidrule(lr){2-5}
                                 & 28                & glass4           & glass          & 4                \\\cmidrule(lr){2-5}
                                 & 29                & glass5           & glass          & 5                \\\cmidrule(lr){1-5}
\multirow{5}{*}{ball}            & 30                & ball1            & ball           & 1                \\\cmidrule(lr){2-5}
                                 & 31                & ball2            & ball           & 2                \\\cmidrule(lr){2-5}
                                 & 32                & ball3            & ball           & 3                \\\cmidrule(lr){2-5}
                                 & 33                & ball4            & ball           & 4                \\\cmidrule(lr){2-5}
                                 & 34                & ball5            & ball           & 5                \\\cmidrule(lr){1-5}
\multirow{5}{*}{marker}          & 35                & marker1          & marker         & 1                \\\cmidrule(lr){2-5}
                                 & 36                & marker2          & marker         & 2                \\\cmidrule(lr){2-5}
                                 & 37                & marker3          & marker         & 3                \\\cmidrule(lr){2-5}
                                 & 38                & marker4          & marker         & 4                \\\cmidrule(lr){2-5}
                                 & 39                & marker5          & marker         & 5                \\\cmidrule(lr){1-5}
\multirow{5}{*}{cup}             & 40                & cup1             & cup            & 1                \\\cmidrule(lr){2-5}
                                 & 41                & cup2             & cup            & 2                \\\cmidrule(lr){2-5}
                                 & 42                & cup3             & cup            & 3                \\\cmidrule(lr){2-5}
                                 & 43                & cup4             & cup            & 4                \\\cmidrule(lr){2-5}
                                 & 44                & cup5             & cup            & 5                \\\cmidrule(lr){1-5}
\multirow{5}{*}{remote\_control} & 45                & remote\_control1 & remote control & 1                \\\cmidrule(lr){2-5}
                                 & 46                & remote\_control2 & remote control & 2                \\\cmidrule(lr){2-5}
                                 & 47                & remote\_control3 & remote control & 3                \\\cmidrule(lr){2-5}
                                 & 48                & remote\_control4 & remote control & 4                \\\cmidrule(lr){2-5}
                                 & 49                & remote\_control5 & remote control & 5 \\\cmidrule(lr){1-5}             
\end{tabular}
}
\end{table}

\paragraph{Hyperparameters.}
Details of hyperparameters for each experiment are shown in Table \ref{tab:hyperparam}.

\begin{table}[]
\caption{Details of hyperparameters.
\label{tab:hyperparam}}
\centering
\resizebox{1.0\textwidth}{!}{
\begin{tabular}{c cc cc cc cc c}
\cmidrule(lr){1-10}
\multicolumn{3}{c}{Scenario}                                                                     & \multicolumn{6}{c}{CI}                                                                     & CIR                     \\
\cmidrule(lr){1-10}
\multicolumn{3}{c}{Dataset}                                                                      & \multicolumn{2}{c}{CIFAR-100} & \multicolumn{2}{c}{CUB-200} & \multicolumn{2}{c}{ImageNet} & \multirow{2}{*}{CORe50} \\
\cmidrule(lr){1-9}
\multicolumn{3}{c}{Experiment}                                                                   & $NC=5$        & $NC=10$       & $NC=10$      & $NC=20$      & $NC=50$      & $NC=100$      &                         \\
\cmidrule(lr){1-10}
\multirow{5}{*}{Statistic}  & \multicolumn{2}{c}{No. tasks}                                      & 11            & 6             & 11           & 6            & 11           & 6             & 79                      \\
\cmidrule(lr){2-10}
                            & \multirow{2}{*}{No. classes/task}            & Initital task       & 50            & 50            & 100          & 100          & 500          & 500           & 10                      \\
                            \cmidrule(lr){3-10}
                            &                                              & Rest tasks          & 5             & 10            & 10           & 20           & 50           & 100           & 5                       \\
                            \cmidrule(lr){2-10}
                            & \multirow{2}{*}{No. training samples/task}   & Initital task       & 25000         & 25000         & 3000         & 3000         & 650000       & 650000        & 3000                    \\
                            \cmidrule(lr){3-10}
                            &                                              & Rest tasks          & 2500          & 5000          & 300          & 600          & 65000        & 130000        & 1500                    \\
                            \cmidrule(lr){1-10}
\multirow{6}{*}{Classifier} & \multicolumn{2}{c}{No. training epochs/task}                       & 100           & 100           & 150          & 150          & 15           & 15            & 100                     \\
\cmidrule(lr){2-10}
                            & \multirow{2}{*}{Batch size}                  & AlexNet, ResNet32   & 256           & 256           & 64           & 64           & 64           & 64            & 256                     \\
                            \cmidrule(lr){3-10}
                            &                                              & ViT-B/16            & 64            & 64            & 64           & 64           & -           & -            & -                       \\
                            \cmidrule(lr){2-10}
                            & \multicolumn{2}{c}{Optimizer}                                      & SGD           & SGD           & SGD          & SGD          & SGD          & SGD           & SGD                     \\
                            \cmidrule(lr){2-10}
                            & \multicolumn{2}{c}{Learning rate}                                  & 0.001         & 0.001         & 0.0001       & 0.0001       & 0.0001       & 0.0001        & 0.0001                  \\
                            \cmidrule(lr){2-10}
                            & \multicolumn{2}{c}{Weight decay}                                   & 0.0005        & 0.0005        & 0.0005       & 0.0005       & 0.0005       & 0.0005        & 0.0005                  \\
                            \cmidrule(lr){1-10}
\multirow{15}{*}{Diffusion} & \multirow{2}{*}{No. training steps/task}     & Initial task        & 15000         & 15000         & 15000        & 15000        & 150000       & 150000        & 15000                   \\
\cmidrule(lr){3-10}
                            &                                              & Rest tasks          & 15000         & 15000         & 15000        & 15000        & 30000        & 60000         & 15000                   \\
                            \cmidrule(lr){2-10}
                            & \multicolumn{2}{c}{Batch size}                                     & 64            & 64            & 64           & 64           & 64           & 64            & 64                      \\
                            \cmidrule(lr){2-10}
                            & \multicolumn{2}{c}{Mixed precision}                                & fp16          & fp16          & fp16         & fp16         & fp16         & fp16          & fp16                    \\
                            \cmidrule(lr){2-10}
                            & \multicolumn{2}{c}{Drop label rate (Classifier-free guidance)}     & 0.2           & 0.2           & 0.2          & 0.2          & 0.2          & 0.2           & 0.2                     \\
                            \cmidrule(lr){2-10}
                            & \multicolumn{2}{c}{Optimizer}                                      & AdamW         & AdamW         & AdamW        & AdamW        & AdamW        & AdamW         & AdamW                   \\
                            \cmidrule(lr){2-10}
                            & \multicolumn{2}{c}{Learning rate}                                  & 0.0001        & 0.0001        & 0.0001       & 0.0001       & 0.0001       & 0.0001        & 0.0001                  \\
                            \cmidrule(lr){2-10}
                            & \multicolumn{2}{c}{Weight decay}                                   & 0.01          & 0.01          & 0.01         & 0.01         & 0.01         & 0.01          & 0.01                    \\
                            \cmidrule(lr){2-10}
                            & \multicolumn{2}{c}{Gradient clipping}                              & 1.0           & 1.0           & 1.0          & 1.0          & 1.0          & 1.0           & 1.0                     \\
                            \cmidrule(lr){2-10}
                            & \multicolumn{2}{c}{Training scheduler}                             & DDPM          & DDPM          & DDPM         & DDPM         & DDPM         & DDPM          & DDPM                    \\
                            \cmidrule(lr){2-10}
                            & \multicolumn{2}{c}{No. training timesteps}                         & 1000          & 1000          & 1000         & 1000         & 1000         & 1000          & 1000                    \\
                            \cmidrule(lr){2-10}
                            & \multicolumn{2}{c}{Inference scheduler}                            & DDIM          & DDIM          & DDIM         & DDIM         & DDIM         & DDIM          & DDIM                    \\
                            \cmidrule(lr){2-10}
                            & \multicolumn{2}{c}{No. inference timesteps}                        & 100           & 100           & 100          & 100          & 100          & 100           & 100                     \\
                            \cmidrule(lr){2-10}
                            & \multicolumn{2}{c}{Classifier-free guidance sampling weight $(w)$} & 4.0           & 4.0           & 4.0          & 4.0          & 4.0          & 4.0           & 4.0                     \\
                            \cmidrule(lr){2-10}
                            & \multicolumn{2}{c}{No. generated samples/class}                    & 20            & 20            & 20           & 20           & 20           & 20            & 20                      \\
                            \cmidrule(lr){1-10}
\multirow{2}{*}{Class-protype}               & \multicolumn{2}{c}{Learning rate}                                  & 0.01          & 0.01          & 0.01         & 0.01         & 0.01         & 0.01          & 0.01                    \\
\cmidrule(lr){2-10}
                            & \multicolumn{2}{c}{Weight decay}                                   & 0.01          & 0.01          & 0.01         & 0.01         & 0.01         & 0.01          & 0.01      \\
                            \cmidrule(lr){1-10}
\end{tabular}
}
\end{table}

\clearpage  

%
%




\end{document}